\pdfoutput=1

\documentclass[11pt]{article}

\usepackage{acl}

\usepackage{times}
\usepackage{latexsym}

\usepackage{amsmath}
\usepackage{mathtools}
\usepackage{array}

\usepackage[compact]{titlesec}         
\titlespacing{\section}{0pt}{5pt}{5pt}

\usepackage[T1]{fontenc}

\usepackage[utf8]{inputenc}

\usepackage{microtype}

\usepackage{subcaption}

\usepackage{graphicx}

\title{Language acquisition: do children and language models follow similar learning stages?}

\author{Linnea Evanson \\
  \small Meta AI Paris;\\
  \small Laboratoire des systèmes perceptifs\vspace{-0.5em}\\
  \small École normale supérieure\vspace{-0.5em}\\ 
  \small PSL University\\
  \texttt{\small linnea.evanson8@gmail.com} \\\And
  Yair Lakretz\thanks{\hspace{0.5em}Equal Contribution} \vspace{+0.37em}\\
  \small Cognitive Neuroimaging Unit\vspace{-0.5em}\\
  \small CEA, INSERM\vspace{-0.5em}\\
  \small Université Paris-Saclay\vspace{-0.5em}\\
  \small NeuroSpin Center\\
  \texttt{\small yair.lakretz@gmail.com}  \\\And
  Jean-R\'{e}mi King\hspace{-0.05em}\footnotemark[1] \\
  \small Meta AI Paris;\\
  \small Laboratoire des systèmes perceptifs\vspace{-0.5em}\\
  \small École normale supérieure\vspace{-0.5em}\\ 
  \small PSL University \\
  \texttt{\small jeanremi@meta.com} \\}

\begin{document}
\maketitle

\begin{abstract}
During language acquisition, children 
follow a typical sequence of learning stages,
whereby they first learn to categorize phonemes before they develop their lexicon and eventually master increasingly complex syntactic structures. 
However, the computational principles that lead to this learning trajectory remain largely unknown.
To investigate this, we here compare the learning trajectories of deep language models to those of children.  
Specifically, we test whether, during its training, GPT-2 exhibits stages of language acquisition comparable to those observed in children aged between 18 months and 6 years. 
For this, we train 48 GPT-2 models from scratch and evaluate their syntactic and semantic abilities at each training step, using 96 probes curated from the BLiMP, Zorro and BIG-Bench benchmarks. We then compare these evaluations with the behavior of 54 children during language production. 
Our analyses reveal three main findings. First, similarly to children, the language models tend to learn linguistic skills in a systematic order. Second, this learning scheme is parallel: the language tasks that are learned last improve from the very first training steps. Third, some -- but not all -- learning stages are shared between children and these language models.
Overall, these results shed new light on the principles of language acquisition, and highlight important divergences in how humans and modern algorithms learn to process natural language.
\end{abstract}
\section{Introduction}

Language acquisition is marked by a series of successive stages \citep{dupoux2018cognitive, Kuhl2004, Werker2018}. Within their first year of existence, humans infants successively acquire prosody contours \citep{Mehler1988}, phonetic categories \citep{werker1984,kuhl1992,mazuka2011} and frequent words \citep{tincoff1999, Bergelson2012}. They then learn to produce basic syntactic structures (\emph{e.g.} ``The boy sang'' or ``The boy fell''), questions (\emph{e.g.} ``What sound does a cow make?'') and nested syntactic structures (\emph{e.g.} ``The boy that I saw sang''), at approximately 12, 30, and 42 months, respectively \citep{growing_trees}. Even though some children may take slightly longer to learn than others, there is a set order in which children acquire various syntactic structures \cite{friedmann2021stages}.

Our understanding of the entire learning trajectory of children remains very coarse, however. This partly stems from the difficulty of measuring linguistic skills in young children. In babies, experimenters typically measure eye gaze and sucking rate while children process linguistic stimuli, as these reflexive behaviors are known to increase during surprising events. Such ``implicit'' approaches have successfully been used to assess whether non-speaking infants detect linguistic violations \citep{Zamuner2006}, distinguish lexical from grammatical words \citep{Shi1999} or discriminate their native language from a foreign language \cite{Mehler1988, Kuhl2006, Nazzi2000}.
%
In older children, linguistic skills can also be more explicitly measured from spontaneous speech and sentence repetition. For example, a recent study by \citet{growing_trees}, to which we compare our work in this paper, quantified the extent to which 18 month to 6 year-old children produce variably complex syntactic structures. For both of these approaches, however, the measures from children at such early ages can be noisy and fragmented. 

Interestingly, these issues do not apply to modern language models. Deep learning architectures trained to predict words from their proximal contexts have proved immensely effective at learning to process natural language \citep{gpt2,bert}. Unlike humans, these algorithms can be easily probed during training, at any time point and rate, and with unlimited number of test stimuli, without interfering with their language acquisition \citep{sagot,manning,bowman}. Furthermore, high-performing deep nets have been shown to implicitly \cite{bigbench1, bigbench4} or explicitly learn to represent and use syntactic structures \citep{manning}, as well as to use features such as concreteness and lexical class to learn language \cite{Chang2022}. Finally, and importantly, these deep neural networks have recently been shown to represent lexical, syntactic and compositional representations similarly to the adult brain \cite{huth,toneva,Caucheteux,pasquiou2022neural,pasquiou2023information,caucheteux2023evidence}. Evidencing similar learning trajectories in children and language models could thus provide an invaluable framework to better understand the computational principles underlying language acquisition. 

Here, we compare the trajectory of language acquisition between human children and modern language models. We focus on three main questions. First, do these models learn linguistic skills in a systematic order? Second, is this trajectory sequential or parallel? Third, is this trajectory similar to that of children? These hypotheses are illustrated in Figure \ref{fig:hypotheses}.

Specifically, we train 48 GPT-2 architectures \citep{gpt2} from scratch, using a standard next-word prediction objective. We then evaluate, at each training step, their linguistic abilities with 96  semantic and syntactic probes curated from the BLiMP, Zorro and BIG-Bench benchmarks \citep{blimp,babyberta_zorro,srivastava2022imitation}. Finally, we compare a subset of these probes to the behavior of 54 children aged, between 18 months and 6 years \citep{growing_trees}.

\section{Approach}

\subsection{Language models}
We consider two main language models. First, we use a pretrained language model -- GPT-2 -- as provided by HuggingFace \footnote{https://huggingface.co/gpt2} and pretrained on 40\,GB of data \citep{gpt2}. 
Second, we separately train 48 versions of a 12-layer GPT-2 model from scratch. We train each model on WikiText103 \cite{wiki103} with a distinct random seed to set its initial parameters and data-loader. 
Each model is evaluated on all linguistic probes every 100 training steps. 
Further training details are provided in Appendix \ref{app_modeltraining}. 

\subsection{Zero-shot linguistic probes}
Zero-shot linguistic probes are sentences or phrases crafted to evaluate whether a model has learned a particular linguistic skill, without training or fine-tuning the model on that particular skill. In practice, a zero-shot probe consists of comparing the estimated probability of a grammatical sentence with that of a matched ungrammatical sentence. This two-alternative forced-choice approach can be compared to "acceptability judgements", classically used in linguistics \citep{warstadt2019neural}.

We evaluate our models on 96 different linguistic probes, curated from three open source benchmarks, the details of which are presented in Appendix \ref{app_probes}.

Specifically, we compare the probability of each sentence in a grammatical/ungrammatical pair by evaluating the sum of the logarithm of the loss output by the softmax layer:

\begin{equation} \label{eqn1}
    \sum _{i=0}^{n_g} log(f(X_g)_i) < \sum _{j=0}^{n_u} log(f(X_u)_j )
\end{equation}
with $f$ the softmax layer of the language model, $X_g$ and $X_u$ the grammatical and ungrammatical sentences, respectively, and
$n_g$ and $n_u$, the number of tokens in the grammatical and ungrammatical sentences, respectively.

The accuracy of a given probe is the percentage of pairs where the estimated probability of the grammatical sentence is higher than that of the ungrammatical sentence. 

\subsection{Assessing learning trajectory} \label{permut_method}
To evaluate whether the trajectory of language acquisition is shared across models, we rank the probes by their "acquisition time", \emph{i.e.} the number of steps taken by a model to reach 90\% of its final accuracy on a particular probe, for each model independently. We then assess the correlation of ranks between all pairs of the 48 models and take the average of these correlations. To estimate the statistical significance of this average correlation we redo this calculation for all possible model pairs after shuffling the ranks of one of the models in each pair. We repeat this permutation 1,000 times, getting 1,000 values for this shuffled correlation. If in all cases this shuffled correlation is lower than the true average correlation, then the order of acquisition time is shared across models with p $<$ 0.001.

\subsection{Parallel versus Sequential learning}
\begin{figure*}[htb!]
    \centering
    \includegraphics[width=1\textwidth]{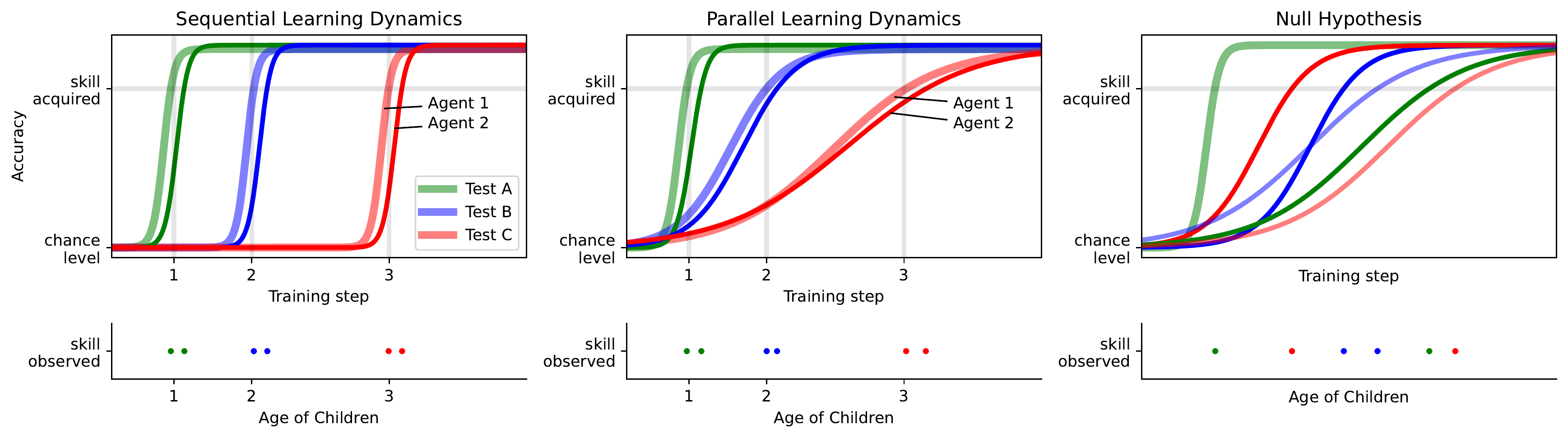}
    \caption{
    Hypotheses. Skill performance (y-axis) as a function of training (x-axis) illustrated on three tasks (colors) and two agents (model or children). Sequential learning implies that the learning of a complex skill (e.g. C, shown in red) does not start before simplest skills (e.g. A in green and B in blue) are fully learned. Parallel learning implies that all skills are acquired simultaneously, but at different speeds. Sequential and parallel learning may cross an arbitrary performance threshold at the same training step.
    The frequency at which we can probe artificial networks during learning is much greater than what is realistically possible in children, giving us the timescale granularity to distinguish sequential and parallel learning trajectories. We also present a null hypothesis, that artificial networks with different random seeds may learn skills in a different order.
    \label{fig:hypotheses}
    }
\end{figure*}

Language acquisition may be characterized by a "sequential" or a "parallel" learning scheme (Figure \ref{fig:hypotheses}).
"Sequential" learning designates the case where a complex skill does not start to be learned before simpler skills are mastered. 
By contrast, "Parallel" learning designates the case where all skills are acquired simultaneously, but at different speeds. 
The null hypothesis is that the order in which an agent learns linguistic skills varies across agents.
To determine the learning scheme of language models, we consider whether the probes have a positive derivative in the first three checkpoints (parallel learning) or not (sequential learning), and whether they have statistically different learning rates (by performing a one-way ANOVA test) across the three groups.

\subsection{Assessing linguistic skill from children's behavior}
\citet{growing_trees} studied 54 Hebrew-speaking children between the ages of 18 - 71 months and investigated the emergence of 11 linguistic phenomena, which the authors propose to organize into three stages (details in Appendix \ref{appen_childrentests}). For our analysis we select the following tests, one from each stage:
\begin{itemize}
    \item Stage 1: Simple sentences in subject-verb (SV) order
    \item Stage 2: Wh Questions
    \item Stage 3: Relative Clauses
\end{itemize}

Data collection consisted of spontaneous speech samples produced by each child at home. Each sample was then manually annotated to detect the presence of each of the linguistic phenomena. A linguistic phenomenon was considered learned if and only if it was present in the speech sample. Speech samples had a mean length of 151 utterances per sample and standard deviation of 37. The aggregated data was made available directly in the original paper (under Creative Commons Attribution 4.0 International License), and here used for comparison with our language models.
In Table \ref{table} we show which probes in the models matched with these tests.

\begin{table*}
\centering
\begin{tabular}{lll}
\hline
\textbf{Stage} & \textbf{Children} & \textbf{Language Model}\\
\hline
1 & Simple sentences in Subject-Verb (SV) order & SV agreement across simple sentences \\
2 & Wh-questions & SV agreement in questions \\
3 & Relative Clauses (RCs) & SV agreement across object RCs \\
\hline
\end{tabular}
\caption{\label{table}
Linguistic phenomena in children selected for comparison with probes in the language models.
}
\end{table*}

\section{Results}

\begin{figure*}[htb!]
    \centering
    \makebox[\textwidth][c]{%
    \includegraphics[width=\textwidth]{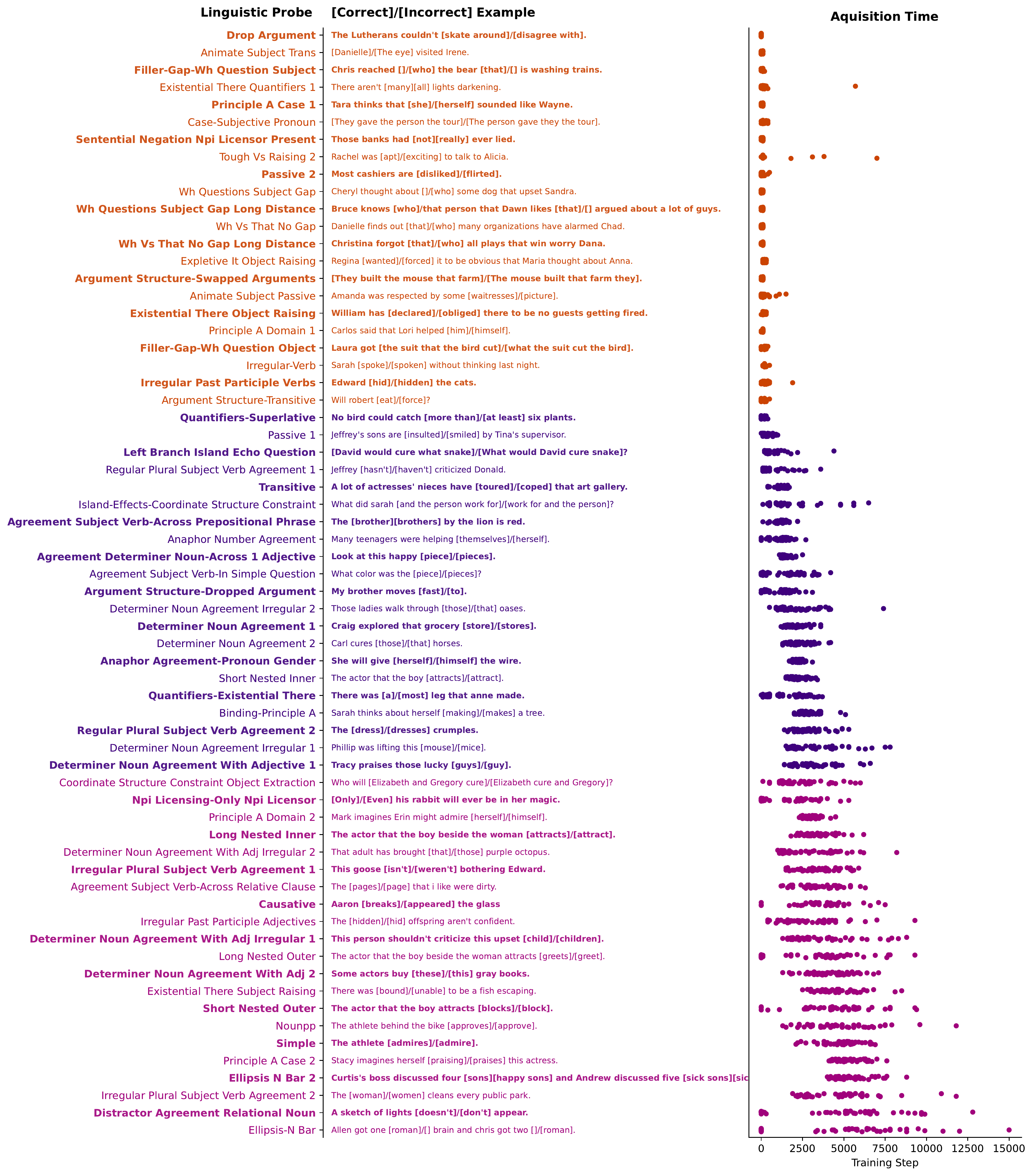}%
    }
    \caption{The performance of the models on each linguistic probe over time is smoothed using a moving average filter with window size of 6 checkpoints then the number of steps required to reach 90\% of final accuracy (acquisition time) is calculated. Probes are ordered by increasing average acquisition time. Results shown for 48 models. Only probes which have final accuracy greater than chance (50\%) are shown. This demonstrates that probes tend to be learned in the same order by all agents with R = 0.743, p < 0.001, disproving the null hypothesis.
    }
    \label{fig:h_bar}
\end{figure*}

We aim to compare the learning trajectories of deep language models to those observed in 54 children aged between 18 months and 6 years. For this, we trained variants of GPT-2 models \cite{gpt2} from 48 different random seeds with the WikiText103 dataset \cite{wiki103} and evaluated each model on 96 linguistic probes every 100 steps. 

At the end of this training, 64 probes (66\%) were achieved above chance level (50\% accuracy) by all models. In comparison, a pretrained version of GPT-2 large \cite{gpt2} provided by Hugging Face\footnote{https://huggingface.co/tftransformers/gpt2-large}, and trained on a much larger dataset\footnote{40 GB compared to the 181 MB of WikiText103}, achieves above-chance performance on 93 of the 96 probes.

\subsection{A systematic learning trajectory} \label{syst_learn_traj}
For clarity, we focus on the learning dynamics of the probes that ultimately achieve above-chance performance in our training. Figure \ref{fig:h_bar} lists all probes learned above chance level, ordered by their average acquisition time. We perform the permutation analysis outlined in \ref{permut_method}, to evaluate whether the order of acquisition is shared between models, and find that their order of acquisition is correlated with $R=0.743$ and p $<$ 0.001. These results suggest that there is a systematic learning trajectory among models.

\subsection{Learning is parallel across linguistic tasks}\label{emperical_3groups}

\begin{figure}[htp]
    \centering
    \includegraphics[width=\linewidth]{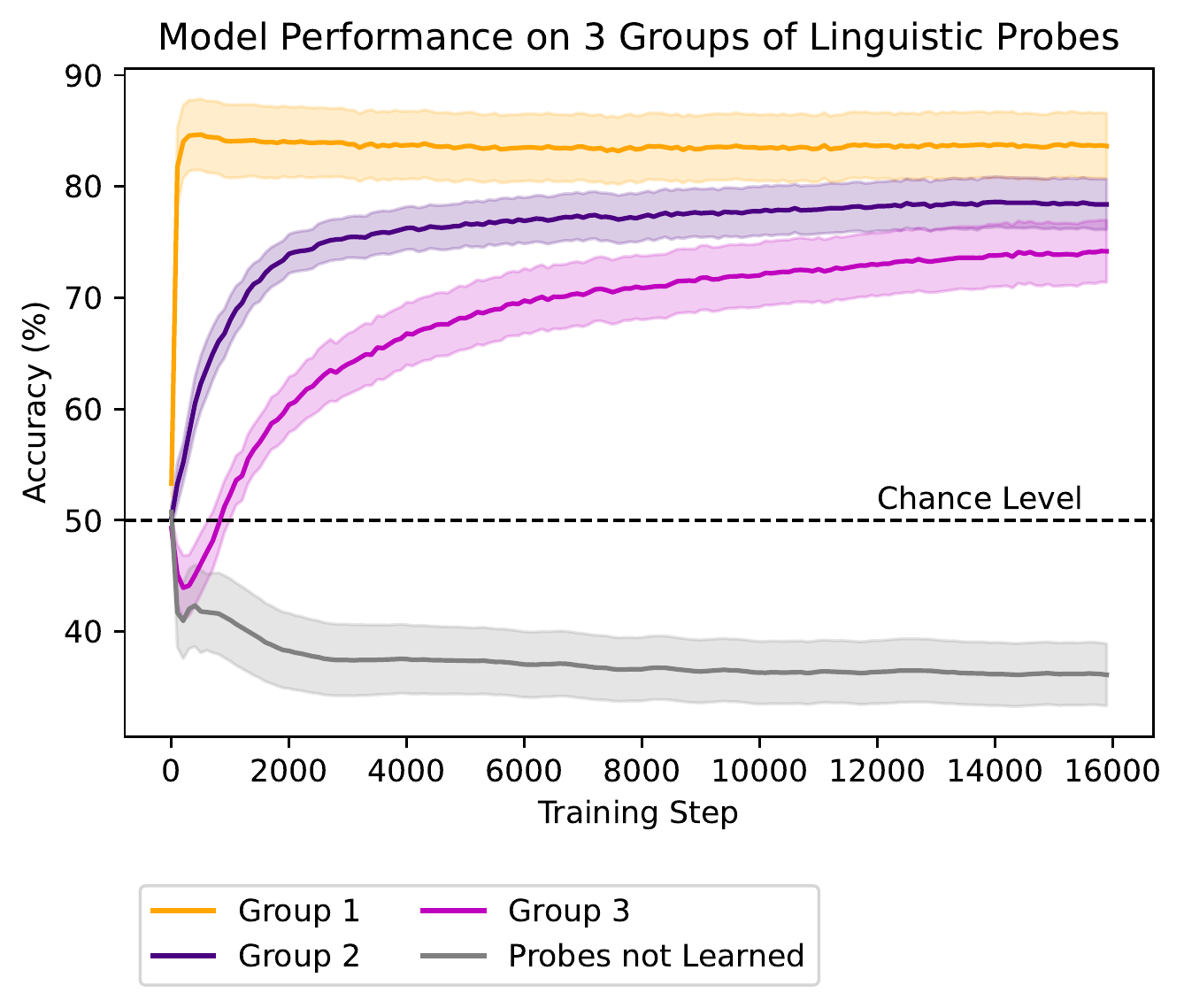}
    \caption{Linguistic probes grouped into 3 groups to avoid plotting one line per probe. The probes in each group correspond to the colours in Figure \ref{fig:h_bar}. Probes which have final accuracy less than chance (50\%) are placed in their own group, and tend to zero likely due to biases towards plural verbs in English (\textit{c.f.} \ref{biases_section}).  Shading is standard error of the mean across probes in the group. This figure demonstrates that linguistic skills are learned in parallel not in sequence.
    }
    \label{fig:3stages}
\end{figure}

Are these linguistic skills learned sequentially or in parallel (Figure \ref{fig:hypotheses})? To address this question, we evaluate whether each linguistic probe starts to improve from the very first training steps but with different rates (i.e. a ``parallel'' learning scheme) or, on the contrary, whether some probes only start to improve once others have reached a particular performance (i.e.  a ``sequential'' learning scheme). As the individual learning trajectories of each probe were noisy, we group the 64 linguistic probes into three categories: early, middle and late acquisition time (Figure \ref{fig:3stages}).   

Overall, we observe parallel learning between the three groups: their performances all increase from the beginning of training: 95\% of tests in all three groups have a positive derivative within the first three hundred steps. However, they have different learning rates, as evaluated with a one-way ANOVA test on the learning rate (i.e. change of accuracy over time) obtained in each group and in each model ($p < 10^{-23}$). 


\subsection{Comparison with children}

\begin{figure*}[htb!]
     \centering
     \makebox[\textwidth][c]{\includegraphics[width=\linewidth]{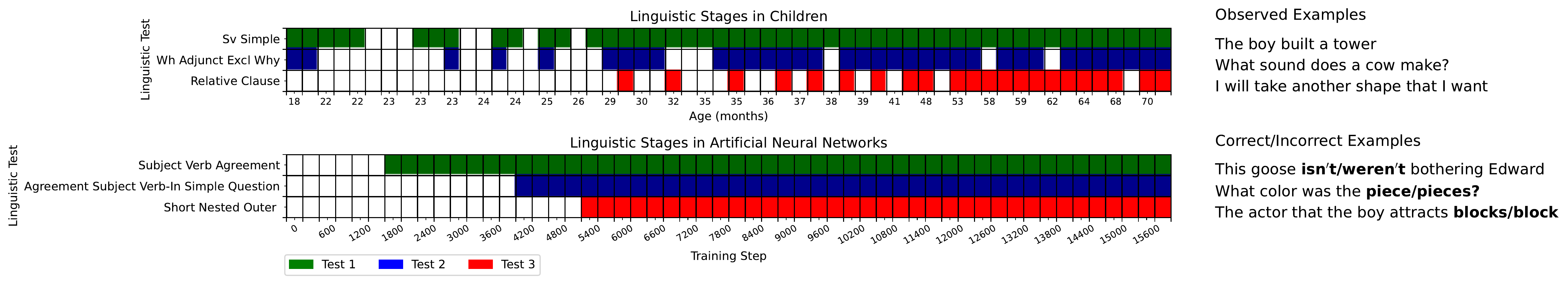}}
    \caption{Comparing language model performance on linguistic probes to children's performance. Example sentences observed in children were originally in Hebrew \cite{growing_trees}. Non-white indicates the phenomena is learned by the agent. The threshold for considering a probe learned by the model is performance above 55\%.}
    \label{fig:matrix}
\end{figure*}

Do these learning trajectories match the behavior of human children? For the three probes that correspond to the three stages identified in children's language acquisition (Table \ref{table}), we observe that the order in which these three probes are learned by the language models is the same as those of children  (Figure \ref{fig:matrix}). This effect is robust across random seeds: 46 of our 48 GPT-2 models follows this order, where chance level is $(1/3!)^{46} = 1.60e^{-36}$. 
For this subset of language abilities, models and children thus seem to acquire syntactic skills in a similar order. 

\subsection{Learning of theoretically-defined stages is parallel}

\begin{figure*}[htb!]
    \centering
    \begin{subfigure}[t]{0.49\textwidth}
        \centering
        \includegraphics[width=\textwidth]{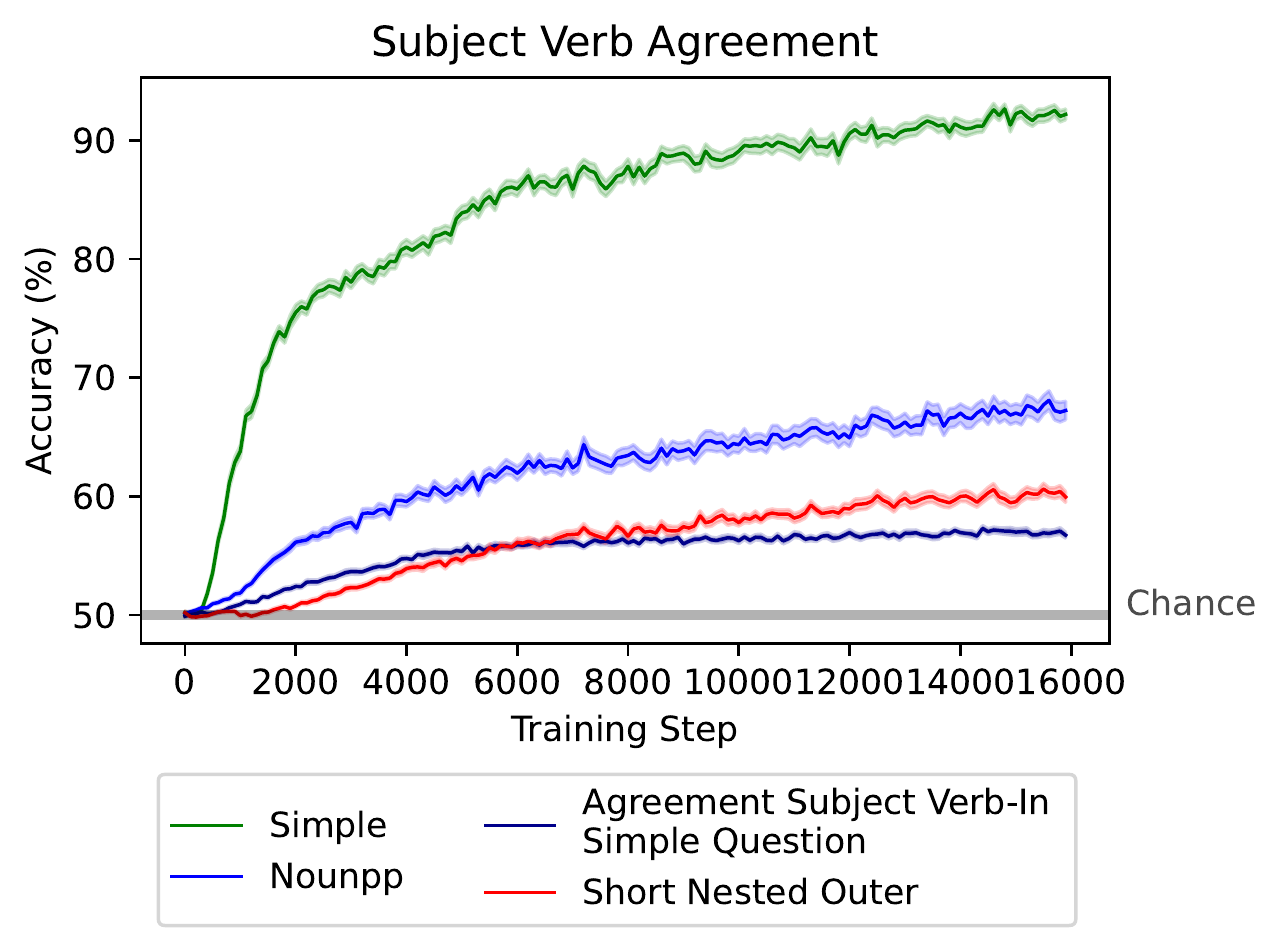}
        \caption{
        }
        \label{fig:deep_dive_avg}
    \end{subfigure}
    \hfill
    \begin{subfigure}[t]{0.49\textwidth}
        \centering
        \includegraphics[width=\textwidth]{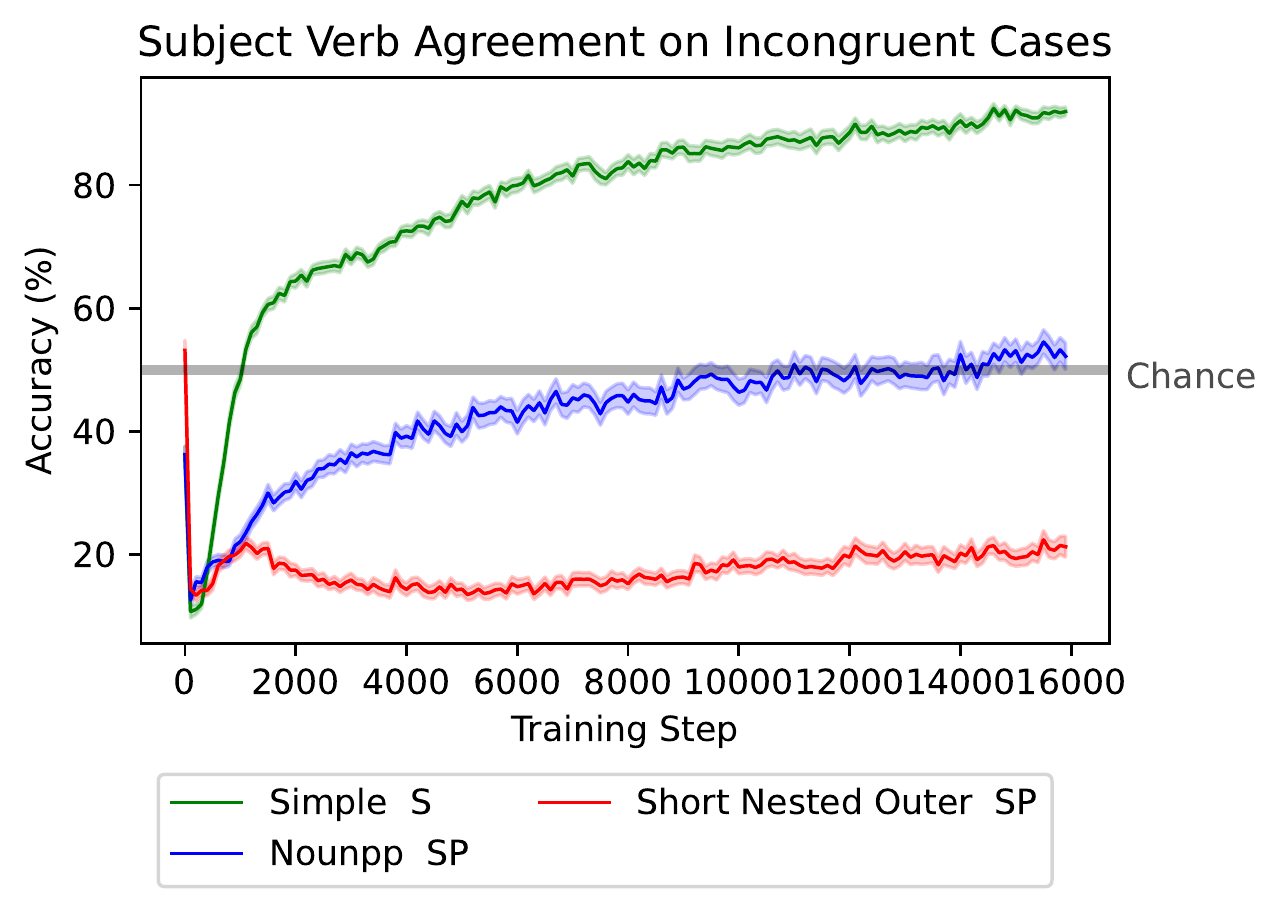}
        \caption{}
        \label{fig:deep_dive}
    \end{subfigure}
    \begin{subfigure}[t]{\textwidth}
         \centering
        \includegraphics[width=\textwidth]{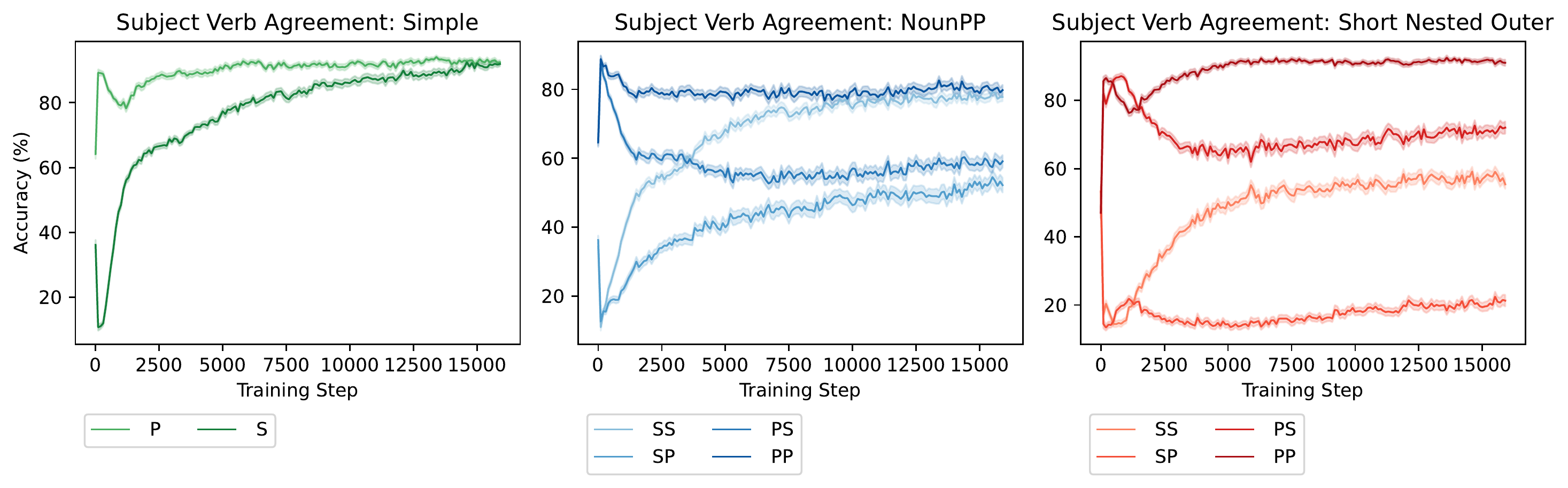}
        \caption{}
        \label{fig:deep_dive_biases}
    \end{subfigure}
    \caption{Subject verb agreement. Shading is standard error of the mean across seeds. S: Singular. P: Plural. \textbf{A}. Parallel learning is observed in the three stages defined by \citet{growing_trees}, when results are averaged across congruent and incongruent cases. \textbf{B}. Subject verb agreement on incongruent cases which indicate whether the model understands syntax. The networks do not learn some syntax structures as the incongruent case of Short Nested Outer does not reach above chance level. \textbf{C}. Development trajectories of the bias towards plural number in English.}
    \label{deepdive}
\end{figure*}

In part \ref{syst_learn_traj}, we showed that GPT-2 learns its language abilities in parallel. Does this learning scheme also characterize the three syntactic skills investigated in children? 
To address this question, we now look at the learning curves of the skills defined in Table \ref{table}, as well as an additional probe: Nounpp, as it can be separated into congruent and incongruent cases which is important for the analysis in section \ref{heuristics}. Overall, we observe that these probes are all learned in parallel in the model (Figure \ref{fig:deep_dive_avg}).

\subsection{Models use both syntax and heuristics} \label{heuristics}
Both an understanding of syntactic rules and a superficial heuristics can lead to above-chance performances on these probes. Indeed, in many sentences (\emph{e.g.} The cat [subject] of the lady [attractor] is [verb] hungry.), the number of the verb is congruent with the number of the adjacent attractor, even if the two are not related syntactically. To verify that the GPT-2 models effectively learn the syntactic rules, we thus separately examine congruent and incongruent cases. Incongruent cases require knowledge of the syntax of the sentence as the correct verb number is different from the number of the attractor. Empirically, we observe that the models do not learn the incongruent case in stage three above chance level, and just barely reach chance level on the incongruent case in stage two (Figure \ref{fig:deep_dive}), indicating that our models are using heuristics rather than syntactic rules to achieve high accuracy on the congruent cases (leading to above chance performance on the probe overall in Figure \ref{fig:deep_dive_avg}). On the contrary, the pretrained GPT-2 large achieves above 75\% accuracy also on the incongruent cases of these probes. Thus for the models trained on the WikiText103, syntax is only learned for stages one and two, and heuristics seem to explain the above chance accuracy in stage three. A larger training dataset is required to learn syntax and not only heuristics for the most difficult examples.

\subsection{Impact of number biases in congruent and incongruent sentences} \label{biases_section}
In previous work, it was found that a variety of language models have a bias towards plural English verbs, and several studies \cite{Jumelet2019, Lakretz, bigbench2} 
determined that LSTM-based models have a default number and gender prediction preference.
To examine whether number bias has a significant effect on our analysis, we compare congruent sentences with only singular or only plural verbs and incongruent sentences with a plural or a singular verb. Accuracy on predicting plural verbs increases sharply from the start of the training and then drops. By contrast, the accuracy of singular cases first drops and then rises (Figure \ref{fig:deep_dive_biases}), indicating that the models are biased towards plural verbs at the beginning of training. This bias is overcome for the stage one probe but for stage two and three it remains throughout training. This explains the initial downward trend in Group 3 and why the unlearned probes tend toward zero in Figure \ref{fig:3stages}.


\section{Discussion}

The stages followed by children to acquire language has been the topic of intense research  \citep{dupoux2018cognitive, Kuhl2004, Werker2018}. While this learning trajectory is becoming clearer for sub-lexical representations \citep{dupoux2018cognitive}, the acquisition of higher-level syntactic and semantic processing remains largely unclear. Here, we approach this long-lasting question through the lens of a deep language architecture, GPT-2 \citep{gpt2}, to test whether this model follows a learning trajectory similar to  children.

\subsection{Language acquisition: similarities and differences between humans and GPT-2}
First, we show that GPT-2 models tend to learn a battery of linguistic phenomena \citep{blimp,bigbench1,babyberta_zorro} in a consistent order. It is the reliability of the acquisition trajectory that allows a direct comparison with the learning trajectory of children \cite{growing_trees}. However, this consistency in GPT-2 models may result from two non-mutually exclusive factors, that remain to be disentangled: either the acquisition time of each linguistic phenomenon relates to its frequency in natural language (e.g. Simple subject-verb-complement are more frequent in natural language than nested syntactic structures; \citealt{karlsson2007constraints}), and/or it relates to their intrinsic complexity (e.g. sentences with nested structure require more operations to be composed than simple sentences). Future work systematically controlling for these relative frequencies is thus necessary to distinguish these two possibilities, and would build upon work by \citet{weber-etal-2021-language} who found that less frequent linguistic phenomena can be learned from fewer examples, though later in training. 

Second, we show that the order in which linguistic skills are acquired is similar between children and GPT-2 -- at least on the syntactic phenomena that were evaluated in these two cohorts, and with the limitation of using number agreement as a proxy to whether the models acquire the corresponding syntactic structure. Similarly to children, GPT-2 models master subject-verb agreement in SV sentences before they master it in questions, or across nested center-embedded clauses (object-relative clauses). This result thus complements a series of studies comparing modern language models and humans. For example, a recent study showed that transformers trained on child-directed data can achieve comparable accuracy on linguistic probes to large pre-trained models \cite{babyberta_zorro}. Similarly, several studies have recently shown that the representations of GPT-2 become increasingly similar to those of the adult human brain during its training \cite{Caucheteux}. Finally, \citet{lavechin2022statistical} showed that models trained on audio in a self-supervised fashion learn phoneme and lexical abilities in a similar trajectory to children.

\subsection{A work in progress}

It is important to stress that significant work remains to be done before drawing any definitive conclusions about the similarities between language acquisition in humans and algorithms. 

First, we only consider a single architecture (GPT-2, \citet{gpt2}) with a unique textual corpus (WikiText103). Testing whether our results hold true independently of the model and training corpus remains an important milestone for future research.

Second, linguistic abilities are not tested with the same protocols in children and in the models: the models are explicitly tested on next word prediction, with a two-alternative forced-choice metric, whereas children were implicitly evaluated on their ability to spontaneously use specific syntactic structures during natural speech. 

Third, there were only three linguistic features that were directly comparable between the model probes and the data in children, and all were syntactic. This leaves a significant margin of progress to modulate our conclusion, and investigate whether the lexicon, narrative structures, pragmatics and world-knowledge are acquired in the same order in humans and algorithms. 

Fourth, the order in which some linguistic skills were learned by GPT-2 does not trivially fit with linguistic theory. For example, the probe ``Simple'', which examines subject-verb agreement in a simple sentence, was one of the last probes to be learned by GPT-2 (it is part of group three in Figure \ref{fig:h_bar}). By contrast, "Wh Questions Subject Gap Long Distance" was among the first probes to be to be learned, even though it would be expected to be much harder than ``Simple''. This unexpected result may be due to the way we approximate "Acquisition Time", namely, the moment when the probes reaches 90\% of the final accuracy. Consequently, probes with very low final accuracy could end up with a shorter Acquisition Time, because noise may lead to crossing the 90\% threshold relatively quickly.

Finally, we show that our models appear to use heuristics rather than a deep understanding of syntax for the most difficult linguistic probes (incongruent numbers between verbs and their attractors) and were biased towards plural English verbs. While our models learn only 66\% of tasks to above chance level, a larger GPT-2 pretrained on considerably more texts successfully perform on 97\% of the tasks, and has an accuracy above 75\% on the incongruent examples, meaning this bias and reliance on heuristics could potentially be solved by training on a larger dataset. 


In sum, additional work remains necessary to identify the exact elements of convergence and divergence between the acquisition of language in models and in children.

\subsection{Fueling the debate between nativism versus empiricism}
The present study fuels a long-lasting debate on the acquisition of language. While ``empiricists`` argue that language can be acquired with a statistical approach
\citep{clark2002unsupervised, kolodny2015learning, chater2018language, mccauley2019language}, ``nativists`` maintain that this ability depends on a core and innate operation, specific to humans \citep{chomsky1959verbal, chomsky1971problems}.

The present study shows how  modern language models may contribute to resolving this debate, by systematically studying which components of a model (e.g. architecture) or properties of the training data (e.g., frequency of sentence structures) contribute to shape the trajectory of language acquisition. 
Claims about an innate Universal Grammar could be understood as an inductive bias of a language model, implemented in its architecture and dynamics, which tightly constrains learning trajectories across models. If this bias is hierarchical (rather than linear) then this could lead to learning trajectories that follow the structure of the syntactic tree, consistently with the hypothesis of three linguistic stages presented by \citet{growing_trees} in humans and what we find in this study in language models. Indeed, neural language models have been previously shown to have a weak inductive bias towards hierarchical processing \cite{mccoy2020does, kharitonov2020they}, which can partially explain our results.

This result echos the recent observation that syntactic trees spontaneously emerge in the middle layers of neural language models \citep{hewitt2019structural}. 
Together, these elements thus suggest that modern neural networks provide fruitful models of language acquisition and could reconcile or settle the confronting theories of language acquisition \citep{warstadt2022artificial}.

\subsection{Conclusion}

Overall, the similarities identified between children and GPT-2 suggest that there may be a small set of means by which to efficiently acquire language. This result is anything but trivial: humans and deep neural networks have extraordinarily different architectures, training, and language exposure. If generalized, this systematic learning trajectory would support the existence of an intrinsic hierarchy of linguistic structures that both machines and humans must climb, be that through inductive biases or properties of the training data, to master the faculty of language. And while these hypotheses remain open, the path to resolve them has never been clearer.

\newpage
\section*{Acknowledgements}
We would like to thank Dieuwke Hupkes, Naama Friedmann, Marco Baroni and the attendees of the EviL meetings for their comments and suggestions.

This project has received funding from the European Union’s Horizon 2020 research and innovation program under the Marie
Skłodowska-Curie grant agreement No 945304, for L.E for her work at PSL. 
This work was funded in part by FrontCog grant ANR-17-EURE-0017 for the work of L.E. and J.R.K. for their work at PSL.

\bibliography{custom}
\bibliographystyle{acl_natbib}

\appendix

\section*{Appendix}
\section{Tests in Children}\label{appen_childrentests}

Detailed description of tests available in children, in the three linguistic stages defined by \cite{growing_trees}:
\begin{itemize}
    \item Stage 1: Subject-Verb Simple, Subject-Verb Unaccusative, Verb-Subject Unaccusative
    \item Stage 2: Root WH-Argument, WH-Adjunct Excluding Why, Preposed Adverb, Root y/n
    \item Stage 3: Why, Relative Clause, Topicalisation, Embedding
\end{itemize}

The probes chosen for comparison (stated in Table \ref{table}), were the only probes that matched well with one of the test available in children. In addition Nounpp was examined in the models, as it fits into the linguistic stage 2, and, as it is part of the BIG-Bench probes, could be separated into congruent and incongruent sentences.

\section{Model Training}\label{app_modeltraining}

To evaluate linguistic abilities of a high-performance language model, we first use the HuggingFace pretrained GPT-2 large which has 774M parameters and is trained on 40GB of data. This model has one-shot perplexity of 22 on WikiText103 \cite{gpt2}.

Then, to evaluate how linguistic abilities vary with language acquisition, we separately trained 48 models (each with a distinct random seed which set the model's initial parameters and the seed of the dataloader) using the 12-layer GPT-2 architecture \cite{gpt2} provided by HuggingFace\footnote{https://huggingface.co/gpt2} on WikiText103 \cite{wiki103} with a learning rate of $10^{-5}$ and a batch size of 16 distributed over 8 GPUs, making a total batch size of 64 and a context size of 128. Training was stopped when then validation loss plateaued, reaching final perplexity of 28 after 10 epochs. This is lower perplexity than the one-shot performance of the HuggingFace pretrained 12-layer GPT-2 which was 37.5, which is logical as our model was trained specifically on this dataset. 

In all cases we used the pretrained tokenizer which has vocabulary size of 50,257. All other parameters were the default training arguments for the transformer provided by HuggingFace. The HuggingFace architectures are publicly available under an MIT license, and WikiText103 is available under Creative Commons Attribution-ShareAlike License.

\section{Linguistic Probe Benchmarks}\label{app_probes}

We use three different zero-shot benchmarks. The first benchmark, `BLiMP' (The Benchmark of Linguistic Minimal Pairs for English) \cite{blimp} contains 67 different probes, each in the form of 1,000 pairs of grammatical and ungrammatical sentences designed to isolate a specific linguistic phenomenon. Adult human performance on BLiMP is 96.4\% \cite{blimp}. The second benchmark, `Zorro' \footnote{https://github.com/phueb/Zorro}, was developed with a vocabulary frequent in child-directed corpora. Zorro contains 13 probes, each consisting of 2,000 pairs of sentences. Finally, the third benchmark is the Subject-Verb Agreement Task of BIG-Bench \cite{srivastava2022imitation, bigbench1, bigbench2, bigbench3, bigbench4}. We focus on the syntactic probes, namely: "Simple English" which contains 600 pairs, "NounPP" which contains 2,400 pairs, and "Short Nested Inner", "Short Nested Outer", "Long Nested Inner" and "Long Nested Outer" which each contain 4,096 pairs of grammatical and ungrammatical sentences. 

Accuracy on a linguistic probe is evaluated with the Jiant \cite{jiant} and UnMasked method \cite{unmasked}. In practice, sentences are input to the model in batches of 300, with padding on the left to make all sentences the length of the longest sentence in the batch. The logit values of punctuation are discarded when estimating the probability of a sentence.

Zorro, Jiant and UnMasked are publicly available under the MIT License, BLiMP under a CC BY License, and BIG-Bench under the Apache License 2.0.

\end{document}